\documentclass[letterpaper, 10 pt, conference]{ieeeconf}  % Comment this line out if you need a4paper

\IEEEoverridecommandlockouts                              % This command is only needed if 
\usepackage{graphics} % for pdf, bitmapped graphics files
\usepackage{graphicx}
\usepackage{amsmath} % assumes amsmath package installed
\usepackage{amssymb}  % assumes amsmath package installed
\usepackage[ruled]{algorithm2e}
\usepackage{booktabs}
\usepackage{multirow}
\usepackage{multicol}
\usepackage{cite}

\title{\LARGE \bf
Self-Supervised Disentangled Representation Learning for\\ Third-Person Imitation Learning
}

\author{Jinghuan Shang$^{1}$ and Michael S. Ryoo$^{1}$% <-this % stops a space
\thanks{*This work was supported by Institute of Information \& communications Technology Planning \& Evaluation (IITP) grant funded by the Ministry of Science and ICT (No.2018-0-00205, Development of Core Technology of Robot Task-Intelligence for Improvement of Labor Condition). This work was also supported by the National Science Foundation (IIS-2104404).}
\thanks{$^{1}$All authors are with the Department of Computer Science,
        Stony Brook University, 100 Nicolls Road, Stony Brook, NY 11794, USA.
        {\tt\small \{jishang, mryoo\}@cs.stonybrook.edu}}%
}

\begin{document}

\maketitle
\thispagestyle{empty}
\pagestyle{empty}

%%%%%%%%%%%%%%%%%%%%%%%%%%%%%%%%%%%%%%%%%%%%%%%%%%%%%%%%%%%%%%%%%%%%%%%%%%%%%%%%
\begin{abstract}
Humans learn to imitate by observing others. However, robot imitation learning generally requires expert demonstrations in the first-person view (FPV). Collecting such FPV videos for every robot could be very expensive.

Third-person imitation learning (TPIL) is the concept of learning action policies by observing other agents in a third-person view (TPV), similar to what humans do. This ultimately allows utilizing human and robot demonstration videos in TPV from many different data sources, for the policy learning. In this paper, we present a TPIL approach for robot tasks with \emph{egomotion}. Although many robot tasks with ground/aerial mobility often involve actions with camera egomotion, study on TPIL for such tasks has been limited. Here, FPV and TPV observations are visually very different; FPV shows egomotion while the agent appearance is only observable in TPV. To enable better state learning for TPIL, we propose our disentangled representation learning method. We use a dual auto-encoder structure plus representation permutation loss and time-contrastive loss to ensure the state and viewpoint representations are well disentangled. Our experiments show the effectiveness of our approach.

\end{abstract}
%%%%%%%%%%%%%%%%%%%%%%%%%%%%%%%%%%%%%%%%%%%%%%%%%%%%%%%%%%%%%%%%%%%%%%%%%%%%%%%%
\section{INTRODUCTION}
Humans learn to imitate by observing others.
In robotics, imitation learning enables a robot to learn complex tasks with minimal environmental knowledge~\cite{hussein2017imitation} based on expert demonstrations. The robot learns a mapping from states (observations) to actions by using the expert trajectories as training data.
However, imitation learning is known to be costly in collecting such expert data.
Such expert demonstrations should be in the first-person view (FPV) from the same (or very similar) viewpoint to the robot and should include actual action labels.
Collecting a sufficient amount of robot data could potentially be very expensive, especially for action labels.

Third-person imitation learning (TPIL) is the concept of learning action policies by observing other agents in a third-person view (TPV) without accessing the action labels.
TPIL allows utilizing many human or robot demonstration videos in TPV from different data sources and very different viewpoints.
It is similar to that humans could map TPV observations to their egocentric perspective~\cite{premack1978does,meltzoff1988imitation} and learn from them. 
Recent advances in TPIL\cite{Stadie2017ThirdPersonIL, sermanet2018time, mees2020adversarial} solved tasks by learning a joint visual state representation space shared by FPV and TPV from synchronized FPV and TPV demonstrations.
Such joint state representation can be used to guide downstream policy learning.

\begin{figure}[tbp]
    \centering
    \includegraphics[scale=0.41]{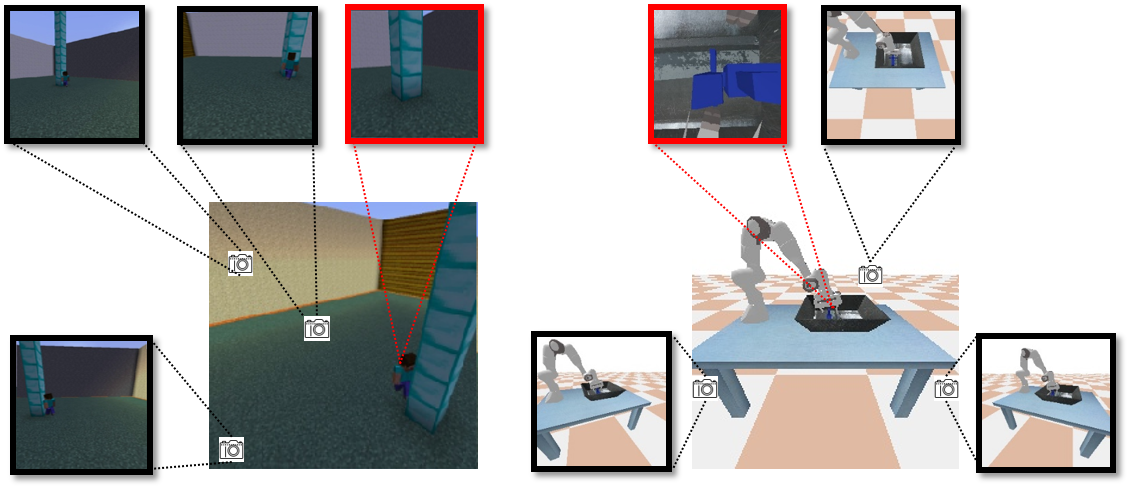}
    \caption{Examples of third-person imitation tasks using Minecraft \textbf{(a)} and Panda robot sim \textbf{(b)}. 
    Red box: FPV that the agent actually observes, displaying agent's egomotion. Black box: TPVs that are taken from multiple fixed cameras. In Minecraft environment, the player in Minecraft is trying to walk to the diamond blocks (blue blocks). In Panda environment, the Panda robot is trying to pick up the object in the tray, based on the observations from a camera mounted on the end effector. }
    \label{fig:egofpv}
    % \vspace{-0.1in}
\end{figure}

In this research, we study a TPIL case where the robot's FPV contains \emph{egomotion} caused by actions (Figure~\ref{fig:egofpv}).
For example, many robot tasks with ground (or aerial) mobility often involve actions with egomotion.
Previous research~\cite{Stadie2017ThirdPersonIL, sermanet2018time} only focused on FPV observations from a static camera, without considering tasks with the egomotion.
Learning a joint visual state representation in such setting is very challenging due to the visual differences between egocentric FPV and TPV: (1) FPV videos consist of agent's egomotion while TPV videos consist of a relatively fixed scene with the moving agent, and (2) the agent itself is not visible (or only a very small portion of it is visible) in the FPV videos.
The study on TPIL for egocentric videos will further broaden the scope of where TPIL algorithms could be applied.

To this end, we introduce a TPIL approach that learns disentangled state and viewpoint representations to ensure state representations are viewpoint-agnostic.
Focusing on such visual differences between FPV and TPV, we propose a dual auto-encoder model to process the FPV and TPV inputs separately.
In addition, we split the latent representation $z$ of each auto-encoder into two parts, $h$ and $v$, to encode disentangled state and viewpoint information, respectively.
We propose representation permutation loss to train $h$ and $v$ to be well-disentangled representations.
We adapt time-contrastive loss and apply it on $h$ to encode state information.
A general reconstruction loss is used to fully train our decoders.
% Such decoders will ensure representation permutation loss is accurate.

Our visual representation learning is based on synchronized FPV and TPV videos from multiple viewpoints. Once finished, our policy learning is done by providing a single TPV expert demonstration without any FPV experts. 

\section{RELATED WORK}
Our work is mainly related to three categories of research: imitation learning from observation, disentanglement, and contrastive learning.
\subsection{Imitation learning from observations}
Imitation learning from observations extends imitation learning to the case that experts' action labels are no longer available.
Edwards et al.~\cite{edwards2018imitating} first learns a latent policy and then remaps outputs from latent policy to actual action space.

In \cite{liu2018imitationfrom,yang2020cross}, a context translation model is used to map across contexts considering context differences between demonstration and agent's observation.

Third-person imitation learning further extends the imitation learning from observations by replacing FPV expert demonstrations with TPV ones.
Recent literature on third-person imitation learning focuses on how to make connections between TPV and FPV observations by learning visual representations.
TPIL\cite{Stadie2017ThirdPersonIL} extends GAIL\cite{ho2016generative} and uses a domain confusion constraint to force features from both FPV and TPV are indistinguishable for a discriminator. TCN\cite{sermanet2018time} uses a time-contrastive way to learn representations by self-supervised metric learning. TCN~\cite{sermanet2018time} is also extended to other tasks such as skill transfer\cite{mees2020adversarial} and playing hard exploration games\cite{aytar2018playing}. 
mfTCN\cite{dwibedi2018learning} extends the time-contrastive method to multiple frames.
In this work, we extend TPIL to an egocentric FPV setting.

% disentanglements
\subsection{Disentanglement}
Disentanglement is learning independent attributes encoded into separated dimensions of representation space.
Typical methods for disentanglement are based on variational auto-encoders (VAE)\cite{kingma2013auto} and generative-adversarial networks (GAN)\cite{goodfellow2014generative}.

$\beta$-TCVAE\cite{chen2018isolating} hypothesis that each dimension in representation $z$ is mutually independent.
Both CVAE\cite{sohn2015learning} and Info-GAN\cite{chen2016infogan} provide the model with class label input to learn representations that are independent of these labels.
Also, GAN methods like style-GAN\cite{karras2019astyle} get a latent representation as a prior from a normal distribution.
CC-VAE\cite{nair2020contextual} adapts CVAE\cite{sohn2015learning} to a self-supervised manner using an extra auto-encoder to learn a condition representation.

In imitation learning context, representations could also be implicitly disentangled. 
TRAIL\cite{zolna2019task} extends GAIL\cite{ho2016generative} to disentangle task-relevant and task-irrelevant information by adding a consistency constraint on samples in an invariant set. 
The learned representations are agnostic to task-irrelevant factors like colors.
Disentangled representations have been also applied to other computer vision tasks and get successful results \cite{peng2019domain, jiang2019disentangled,shu2017NeuralFace,zhang2019gait,denton2017unsupervised}.

Compared with the above methods, our method explicitly splits latent representation into two components: state and viewpoint representation.
We do not assume that exact viewpoint coordinates (i.e. labels of viewpoints) are provided for disentanglement, and rather train the model in a self-supervised manner.

% contrastive learning and representation learning
\subsection{Contrastive learning}
Contrastive learning performs a comparison across different views that are generated from one single input to learn representations in a self-supervised fashion\cite{aytar2017cross,doersch2015unsupervised}.
The concept of contrastive learning also has been adopted for TPIL~\cite{sermanet2018time}.
For computer vision tasks, CMC\cite{tian2020cmc} takes advantage of InfoNCE loss~\cite{aytar2017cross} on the comparison between anchor sample, positive sample, and a batch of negative samples.
Other research on contrastive learning\cite{chen2020simple,chen2020big,he2020momentum, grill2020bootstrap, chen2020exploring} focus on avoiding trivial solutions to improve representation quality and increasing training efficiency by getting rid of negative samples.
Contrastive learning concepts and techniques are also applied to control tasks like \cite{srinivas2020curl}.
We extend time-contrastive loss~\cite{sermanet2018time} (which is based on triplet loss~\cite{hermans2017defense}) using InfoNCE loss~\cite{aytar2017cross} that includes a batch of negative samples.
We also take advantage of the stop-gradient technique from the above literature to avoid trivial solutions.

\section{PROBLEM SETUP}
\begin{figure*}[thbp]
    \centering
    \includegraphics[scale=0.42]{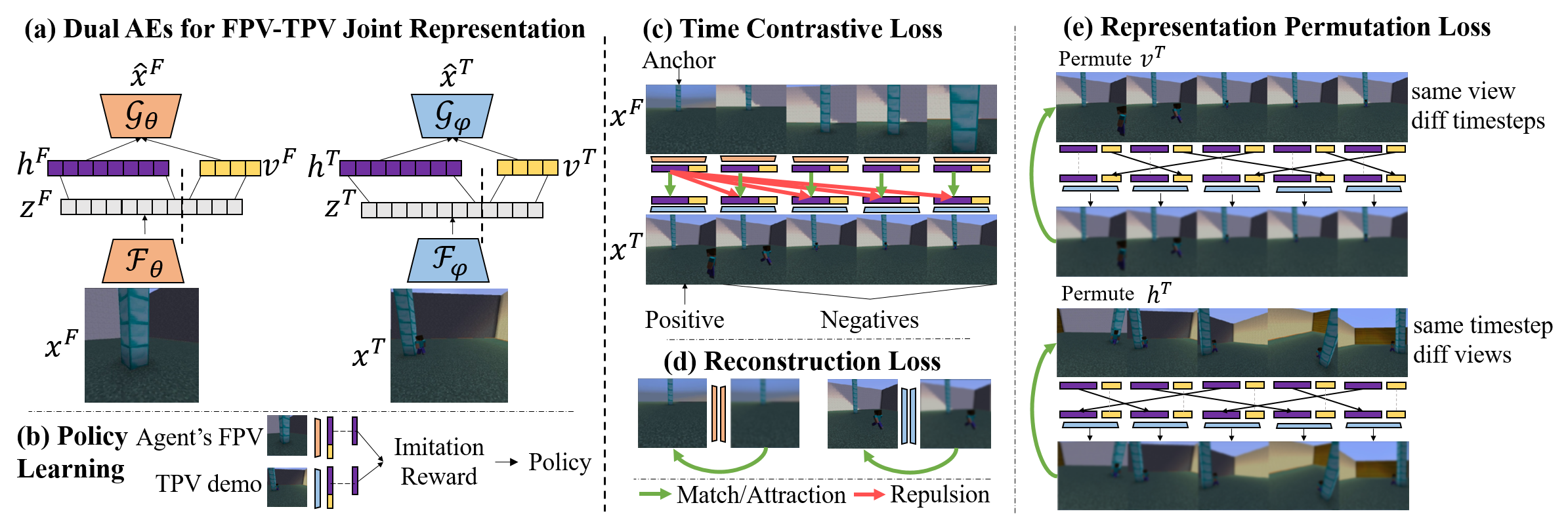}
    \caption{This figure includes an overview of the dual AE model we propose, the policy learning method we use after having a state representation space, and losses we use to learn our FPV-TPV joint state representation. (a) Proposed dual AE model. (b) When doing policy learning, we discard $v$ and use $h$. (c) Time-contrastive loss requires an anchor sample, a positive sample, and a batch of negative samples. This figure uses anchor FPV frame as an example. Symmetrically, a TPV frame can be an anchor. (d) We reconstruct every input FPV/TPV frame and calculate the reconstruction loss. (e) We permute viewpoint representation $v$ among a batch of frames from the same TPV but different timesteps. Symmetrically, we permute state representations $h$ among a batch of frames from the same timestep but different TPVs. We compare the reconstructed frames with original inputs accordingly to have our representation permutation loss. }
    \label{fig:main}
    \vspace{-0.2in}
\end{figure*}
A discrete-time finite-horizon discounted Markov decision process (MDP) can be noted by $M=(\mathcal{S}, \mathcal{A}, \mathcal{P}, r, T)$, where $\mathcal{S}$ is a state set, $\mathcal{A}$ is an action set, $\mathcal{P}$ is a transition probability distribution, $r$ is a reward function, and $T$ is the horizon.
For imitation learning, the agent is given trajectories $\tau=\{(s_t, a_t)\}$ which are states and actions generated from an unknown expert policy $\pi_E$ and $r$ is not traceable.
The agent is required to learn a policy $\pi_\theta$ that recovers $\pi_E$. 

In visual third-person imitation learning from observation, we consider an observer is learning by solely watching a demonstrator's demonstration from a different viewpoint than the demonstrator's, and $a_t$ is not traceable.
Under this setting, the observer and the demonstrator have visual observations $o^T_t\in \mathcal{O}^T$, $\mathcal{O}^T \subseteq \mathbb{R}^{H\times W\times d}$ and $o^F_t \in \mathcal{O}^F$, $\mathcal{O}^F \subseteq \mathbb{R}^{H\times W\times d}$ respectively, where $d=3$ if we consider a pure RGB image input.
Usually, we call $\mathcal{O^T}$ third-person view (TPV) and $\mathcal{O}^F$ first-person view (FPV). 
A fact holds that at any moment $t$, $o^T_t$ and $o^F_t$ are different since they see from different views but $o^T_t$ and $o^F_t$ correspond to a same state $s_t$.
Therefore, visual representation learning approaches explored by ~\cite{Stadie2017ThirdPersonIL,sermanet2018time} tried to solve this problem by finding a function $\mathcal{F}:\mathcal{O} \to \mathcal{H}$ that learns a latent state representation $h=\mathcal{F}(o)$ satisfying $\mathcal{F}(o^T_t)=\mathcal{F}(o^F_t)$ to estimate the true state space $\mathcal{S}$.

In this research, we investigate an egocentric FPV setting that actions cause agent's egomotion. Moreover, we consider TPV demonstrations from multiple different viewpoints. This is more challenging than learning the state embedding only applicable for one third-person viewpoint.

Formally, we have a set of synchronized FPV-TPV trajectories $\{\tau\}$ for visual representation learning, and a single TPV expert demonstration $\tau_E=\{o^{T}_t\}$ generated by some unknown expert policy $\pi_E$ for policy learning. 
For each FPV-TPV trajectory, it has one FPV video and multiple TPV videos from multiple viewpoints: $\tau=(\{o^F_t\}, \{o^{T_1}_t\}, \dots, \{o^{T_n}_t\})$.
Given the egocentric FPV setting, $o^F$ is visually different from $o^T$ since $o^F$ contains egomotion and $o^T$ contains the agent's appearance which is not in $o^F$.
So the challenge is how we design the map function $\mathcal{F}$, i.e. the representation model, that still preserves $\mathcal{F}(o^T_t)=\mathcal{F}(o^F_t)$.
After getting such viewpoint-agnostic state representation $h$, we train a policy $\pi$ to maximize imitation reward formulated by $\tau_E$ to best recover $\pi_E$.
We follow the previous literature\cite{sermanet2018time} to give a distance-metric based reward function, e.g. $r(t)=-||h^F_t-h^T_t||_2$, to perform policy learning, where $h^F_t$ is current agent's FPV observation and $h^T_t$ is given TPV expert demonstration at the same timestep.

\section{FPV-TPV JOINT STATE REPRESENTATION LEARNING}
Focusing on the visual differences between $\mathcal{O}^F$ and $\mathcal{O}^T$, we propose our dual auto-encoder (AE) model with explicitly disentangled latent representations that learns a joint representation space $\mathcal{H}$.
Then, we introduce our modified time-contrastive loss\cite{sermanet2018time} and representation permutation loss to train our proposed model in a self-supervised fashion.

\subsection{Dual AE for Joint State Representation Learning}
Considering the visual differences between $\mathcal{O}^F$ and $\mathcal{O}^T$, we design two different mapping functions, $\mathcal{F}_\theta:\mathcal{O}^F \to \mathcal{Z}$ and $\mathcal{F}_\varphi:\mathcal{O}^T \to \mathcal{Z}$, instead of learning a single universal function for both $\mathcal{O}^F$ and $\mathcal{O}^T$.
We propose a dual AE structure to model such $\mathcal{F}_\theta$ and $\mathcal{F}_\varphi$ as encoders in AEs along with the decoders $\mathcal{G}_\theta$ and $\mathcal{G}_\varphi$.
The decoder $\mathcal{G}_\theta$ takes the full representation $z^F=[h^F,v^F]$ as input and outputs a reconstructed FPV image $\hat{x^F}=\mathcal{G}_\theta([h^F,v^F])$. Similarly $\mathcal{G}_\varphi$ reconstructs TPV image $\hat{x^T}=\mathcal{G}_\varphi([h^T, v^T])$.

We further design the dual AE to disentangle the representation vectors to better learn the joint embedding, taking advantage of multiple training losses.
Specifically, each of our AE splits full representation vector $z$ into two vectors, $z=[h,v]$, where $[\cdot, \cdot]$ means concatenation (See Figure \ref{fig:main}(a)).
Our motivation is to disentangle state and viewpoint from observation.
If well-disentangled, the state representation is viewpoint-invariant, i.e. a joint embedding for TPV and FPV.
Representation decomposition has been an effective strategy to obtain independent components (representations) in multiple computer vision tasks such as object recognition~\cite{peng2019domain, karras2019astyle, Higgins2017betaVAELB, chen2018isolating}, and our motivation is to extend and take advantage of such concept for TPIL.
In our formulation, $h$ (state) and $v$ (viewpoint) are two independent factors we want to split from raw RGB observations.
Similar to \cite{peng2019domain}, which decomposed class-irrelevant features from class-relevant features for object recognition, our formulation has only two factors to decompose. 
Thus, we separate the latent vector into two parts and ensure that they are disentangled by appropriate loss functions.

The main technical challenge is in learning $\mathcal{F}_\theta$ and $\mathcal{F}_\varphi$, while ensuring (1) $h$ to be an FPV-TPV joint state representation and (2) $h$ and $v$ are well disentangled.
In the below subsections, we describe how the time-contrastive loss and the proposed representation permutation loss enable this.

\subsection{Self-Supervised Disentangled Representation Learning}

\subsubsection{Time-Contrastive Loss}
Time-contrastive\cite{sermanet2018time} was previously proposed to make the state representation follow the temporal order.
The temporal order means a state representation $h_t$ should be closed to $h_{t+1}$ but far from $h_{t+n}$ where $n>1$ in most cases.
In practice, we use a triplet that includes anchor (A), positive (P), and negative (N) samples to construct time-contrastive loss.
This loss requires representations of A and P samples are close (attraction), and representations of A and N samples are distant (repulsion).

Original time-contrastive loss\cite{hermans2017defense, sermanet2018time} is in a triplet-loss form. 
We extend it to include multiple negative samples for each single A and P sample using Info-NCE loss (See Figure~\ref{fig:main}(c)), inspired by recent advances in contrastive learning\cite{tian2020cmc,grill2020bootstrap}.
Given two synchronized FPV-TPV demonstration videos, $\{(x^F_i, x^T_i)\}$, we first identify an A(anchor) at timestep $a$.
Suppose $x^F_a$ is the anchor (A) sample, then the $x^T_a$ is the positive (P) sample.
We then randomly sample a batch of negative (N) timesteps $\{n_i\}$ that appears $\xi$ away.
Then $\{x^T_{n_i}\}$ from a TPV video are N samples of $x^F_a$.
After we get A-P-N samples, we have our modified time-contrastive loss term on FPV anchor:
\begin{equation}\nonumber
    \mathcal{L}_{tc}^{F} = -\log \frac{d(h^F_a, sg(h^T_a))}{ d(h^F_a, sg(h^T_a)) + \sum d(h^F_a, sg(h^T_{n_i}))},
\end{equation}
where $h$ is state representation vector, $d(\cdot, \cdot)$ is a critic metric on $h$ implemented based on cosine similarity between $h$ vectors\cite{tian2020cmc}:
\begin{equation}\nonumber
    d(h_1, h_2) = \exp \left(\frac{h_1 \cdot h_2}{|| h_1 || \cdot || h_2 ||} \cdot \tau \right),
\end{equation}
where $||\cdot||$ is L2-norm, $\tau$ is a scaling coefficient\cite{tian2020cmc}, 
$sg(\cdot)$ is a stop-gradient trick\cite{grill2020bootstrap,chen2020exploring} to avoid trivial solutions.
Symmetrically, if we let the TPV frame $x^T_a$ be anchor, we have $\mathcal{L}_{tc}^{T}$ by making $x^F_a$ as positive sample and $\{x^F_{n_i}\}$ negative samples.

To force $h$ to be a joint representation space for FPV and TPV, we add a general matching loss (See Figure \ref{fig:main}(c) in the middle) for all the $\{(h^F_i, h^T_i)\}$ pairs from the same timestep.
We minimize their L2-distance pairwisely:
\begin{equation}\nonumber
    \mathcal{L}_{match} = \mathbb{E}\left[ ||h^T-sg(h^F)||_2 \right],
\end{equation}
where $sg(\cdot)$ is stop-gradient, viewing the state representations from FPV as references.
Together, the total time-contrastive loss is:
\begin{equation}\nonumber
    \mathcal{L}_{tc} = \mathcal{L}_{tc}^{F} + \mathcal{L}_{tc}^{T} + \mathcal{L}_{match}.
\end{equation}

\subsubsection{Representation Permutation Loss}\label{sec:permute}
We introduce our representation permutation loss below to supervise $h$ and $v$ to be disentangled (See Figure~\ref{fig:main}(e)).
Ideally, $h$ and $v$ should be independent, saying that modifying $v$ only changes the viewpoint information, and modifying $h$ only changes the state information.
Therefore, we permute $h$ and $v$ among a batch of samples and reconstruct images from permuted full representations $z=[h_i, v_j]$. 
We expect the reconstructed images should be similar accordingly after the permutation.
We will describe how we permute and compare reconstructed images accordingly below.

Given a single TPV demonstration, we can get the representations from two arbitrary timesteps: $(h^T_i, v^T_i)= \mathcal{F}(x^T_i)$ and $(h^T_j, v^T_j)=\mathcal{F}(x^T_j)$.
Ideally, we should expect their viewpoint information is the same, so we have $v^T_i =v^T_j$.
Then, by exchanging $v^T_i$ to $v^T_j$ and doing reconstruction, we would expect that the reconstruction image should still remain same as input at timestep $i$: $\mathcal{G}([h^T_i, v^T_j]) = x^T_i$.
Symmetrically, $\mathcal{G}([h^T_j, v^T_i]) = x^T_j$.
We can do these exchanges in a batch of samples from a single TPV demonstration by randomly pairing $v$ to $h$.
Then our viewpoint permutation loss is formulated as an reconstruction error
\begin{equation}\nonumber
    \mathcal{L}_{v} = \mathbb{E}_i\left[||\mathcal{G}([h^T_i, v^T_k])-x^T_i||_2\right], k \neq i,
\end{equation}
where $i$, $k$ are time indices of the selected batch.

Symmetrically, we consider permutations on state representations $h$.
Considering samples from the same timestep but different viewpoints, we expect identical $h$ but different $v$.
Similar as $\mathcal{L}_{v}$ above, we randomly pair $h$ to different $v$ and have this state permutation loss
\begin{equation}\nonumber
    \mathcal{L}_{h} = \mathbb{E}_j\left[||\mathcal{G}([h^{T_k}_i, v^{T_j}_i])-x^{T_j}_i||_2\right], k \neq j,
\end{equation}
where $k$ and $j$ indicate different TPVs.
Combine viewpoint and state permutation losses together, we have our representation permutation loss:
\begin{equation}
    \nonumber
    \mathcal{L}_{permute} = \mathcal{L}_{v} + \mathcal{L}_{h}.
\end{equation}
We do not apply the representation permutation loss above to the latent representation $v^F$ from FPV branch, because the viewpoints can continuously change and are highly correlated with agent's egomotion in its FPV.

\subsubsection{Reconstruction Loss}
We have this reconstruction loss to ensure that (a) $h$ and $v$ are meaningful latent representations instead of trivial solutions and (b) we have a reliable decoder $\mathcal{G}$ that can reconstruct images for computing representation permutation loss above.
We compare a reconstructed image $\mathcal{G}(\mathcal{F}(x))$ with input $x$ (See Figure~\ref{fig:main}(d)):
\begin{equation}
    \mathcal{L}_{recon} = \mathbb{E}_x\left[\sum(\mathcal{G}(\mathcal{F}(x))-x)^2\right].
\end{equation}
We apply this loss to all the samples we pass through our model, including the samples used when calculating time-contrastive loss and representation permutation loss.

In conclusion, our overall loss function to learn an FPV-TPV joint representation is
\begin{equation}
    \mathcal{L} = \alpha \mathcal{L}_{tc} + \beta \mathcal{L}_{permute} + \mathcal{L}_{recon},
\end{equation}
where $\alpha$ and $\beta$ are hyper-parameters to control the weight of time-contrastive loss and invariant loss.

\subsection{Implementation Details}
We implement each of our two AEs by seven convolutional layers and seven deconvolutional layers.
As for splitting representation, we can assign any dimension to $v$.
Specifically, if the dimension of $v$ is 0, it means we do not split our full latent representation, i.e. $h:=z$ and thus $\mathcal{L}_{permute}$ is not applicable.
We use TCN\cite{sermanet2018time} as a baseline model that using one universal encoder for FPV and TPV inputs.
It can be regarded as a 0-dimension-$v$ model.
Since we use $h$ for policy learning, we control all models to have 16-dimension $h$ for a fair comparison between models.
We empirically set $\alpha=1$, $\beta=1$ in all our models except ablation studies on these hyper-parameters.

\subsection{Imitation Learning Formulation}
We follow the common setting of imitation learning in recent literature\cite{sermanet2018time, aytar2018playing}.
Given synchronized FPV-TPV demonstrations, we first train a joint representation model (See Figure~\ref{fig:main}(a)).
Then, given only one extra demonstration $\{x^E_i\}$, we let the agent execute the policy in the environment to maximize the imitation learning reward (See Figure~\ref{fig:main}(b)).
The imitation learning reward at each step $R_t$ is assigned based the cosine similarity
\begin{equation}
    R_t =\left\{
        \begin{aligned}
           1,  & &  \cos(h^F_t, h^E_t)\geq\xi \\
           0,  & &  \cos(h^F_t, h^E_t)<\xi  \\
        \end{aligned},
    \right.
\end{equation}
where $[h^F_t, v^F_t] = \mathcal{F}_\theta (x^F_t)$ are the latent representations of the agent's first-person observation and $h^E_t$ is the state representation from expert demonstration, both are at current timestep $t$. $h^E_t$ will be produced by $\mathcal{F}_\theta(x^E_t)$ if we perform a first-person imitation and by $\mathcal{F}_\varphi(x^E_t)$ if it is third-person imitation learning.
We use a threshold $\xi$ to discrete the imitation reward, following the formulation as~\cite{andrychowicz2018hindsight, aytar2018playing}.

\subsection{Policy Model and Policy Training}
We use a multi-layer perceptron (MLP) as our policy model to map the state representation to agent actions: $\pi:\mathcal{H} \to \mathcal{A}$.
The input layer and two hidden layers contain the same dimension as the input state representation.
For continuous action space, the policy outputs means and standard deviations of Gaussian distributions.
For discrete action space, the policy outputs log-likelihoods of actions.
We use PPO\cite{Schulman2017ProximalPO} to optimize the policy based on the imitation learning reward above.

\section{Experiments}
\subsection{Environments and Tasks}
We develop and use two simulated environments to collect data, train, and evaluate methods (See Figure~\ref{fig:egofpv}).
\subsubsection{Minecraft Environment}
We develop a Minecraft game environment based on Project Malmo\cite{mattew2016malmo} and MineRL\cite{guss2019minerldata}.
The agent is a game player and the task is moving itself to the target position.
The agent observation is an RGB visual input from the default FPV in the game without user interfaces such that $\mathcal{O} \subseteq R^{W\times H \times 3}$.
The action space is discrete, where $|\mathcal{A}| = 6$, including moving forward, backward, left, and right, and turning (along with the camera) towards left and right.
The goal is to reach the target labeled as a pillar of diamond blocks (blue ones).
\subsubsection{Panda Environment}
We develop a simulated continuous control environment by PyBullet\cite{coumans2019}.
The agent is a Panda robot mounted on a desk.
The task is reaching an object in a tray in front of the robot and picking up the object.
The agent observation is an RGB visual input, $\mathcal{O} \subseteq R^{W\times H \times 3}$.
The agent has an action space $\mathcal{A} \subseteq [-0.5, 0.5]^{11}$ which is the force applied to 11 joints.
The camera is mounted at the end effector, i.e. the ``hand" of the Panda robot.
The camera follows the motion of the end effector and emulates an egocentric FPV.

\subsection{Dataset Collection}
We collect synchronized FPV-TPV videos from both environments.
For the Minecraft environment, we collect 8 different demonstration trajectories with randomized target diamond block locations.
Each trajectory contains synchronized 1 FPV video and 8 TPV videos.
Third-person viewpoints are not controlled to be the same, so the Minecraft environment is more challenging to get a well-aligned representation.
For Panda environment, we collect 30 different demonstration trajectories with randomized initial locations of the target object.
The expert policy is driven by an oracle using extra information and inverse kinetics.
Each trajectory contains synchronized 1 FPV video and 9 TPV videos corresponding to 9 different viewpoints.
All distinct trajectories share the 9 viewpoints..
We keep the number of trajectories small to better emulate the realistic setting where the amount of training data is limited.
We split 20\% data for each dataset as test sets.

\subsection{Quantitative Evaluations on Representation Space}
We first investigate the quality of our joint representation model in terms of alignment error\cite{sermanet2018time} between FPV-TPV sequences.
Given a FPV frame $x^F_i$ having time index $i$ and its state representation $h^F_i$, we find its nearest TPV state representation neighbor $h^T_j$ that has time index $j$.
The alignment error is the mean absolute temporal distance of $i$ and $j$ regularized by video frame count $L$ over all indices $i$:
\begin{equation}
    alignment\_error = \mathbb{E}_{i}\left[\frac{|i-j|}{L}\right].
\end{equation}
The lower error means the better quality of imitation reward suggesting a better representation space\cite{sermanet2018time}.

\begin{table}[thbp]
    \centering
    \caption{Alignment error comparison between different representation models}
    \begin{tabular}{ccc}
    \toprule
    \multirow{2}{*}{\textbf{Model}} & \multicolumn{2}{c}{\textbf{Environment}}  \\
                          & Minecraft    & Panda     \\
    \midrule
    Multi-view TCN\cite{sermanet2018time}  & 0.2725 & 0.2861  \\
    Ours 0-d $v$ (i.e. without $v$ and $\mathcal{L}_{permute}$)  & 0.2606  & 0.2036  \\
    Ours 4-d $v$  & 0.2398 &  0.1892 \\
    Ours 8-d $v$  & \textbf{0.2329} &  \textbf{0.1550} \\
    Ours 12-d $v$ & 0.2571 &  0.1999 \\
    \bottomrule
    \end{tabular}
    \label{tab:alignerror}
\end{table}
As shown in Table \ref{tab:alignerror}, our model generally has a lower alignment error than the baseline, indicating that our dual AE model can deal with visually different FPV and TPV inputs.
Specifically, Minecraft results show that our model can better align videos from various unseen viewpoints than the baseline.
And from the comparison between non-zero dimensions of $v$ and 0-d $v$, having a disentangled representation $v$ yields better performance than not having it, suggesting that our disentangled representation helps learn state representations.
We further investigate how the non-zero dimensions of viewpoint representation $v$ affect the representation learning quality.
We try 4, 8, and 12 dimensions of $v$.
Table \ref{tab:alignerror} tells that 8-d has the lowest alignment error compared to 4 and 12.
12-d has worse performance than 4-d (due to the overfitting), but we note that all the variants are better than ours 0-d $v$ baseline.
We confirm that having a non-zero-dimension $v$ helps learn a joint representation, while we should choose a suitable dimension of $v$ empirically to ensure the information in $v$ is well encoded, based on how much information we intend $v$ to encode.
We also learn that increasing the dimension of $v$ is not always beneficial to learn such disentangled representation, which is the overfitting problem.

\begin{table}[t]
    \centering
    \caption{Ablation studies about loss functions on Panda environment based on our model with 8-dimension $v$.}
    \begin{tabular}{cc}
    \toprule
        \textbf{Model based on 8-d $v$} & Alignment Error \\
        \midrule
         $\mathcal{L}_{tc} + \mathcal{L}_{permute} + \mathcal{L}_{recon}$ & \textbf{0.1550} \\
         $\mathcal{L}_{tc} + \mathcal{L}_{vmatch} + \mathcal{L}_{recon}$ & 0.2518 \\
         $\mathcal{L}_{tc} + \mathcal{L}_{recon}$ &   0.1922   \\
         $\mathcal{L}_{tc} + \mathcal{L}_{permute}$ & 0.1748\\
         $\mathcal{L}_{tc}$  &  0.1844\\
         $5\mathcal{L}_{tc} + \mathcal{L}_{permute} + \mathcal{L}_{recon}$ &  0.1695    \\
         $\mathcal{L}_{tc} + 5\mathcal{L}_{permute} + \mathcal{L}_{recon}$ &  0.1616   \\
         \bottomrule
    \end{tabular}
    \label{tab:ablation}
\end{table}
We also do ablation studies about our loss functions on Panda environment (see Table \ref{tab:ablation}).
We first compare our $\mathcal{L}_{permute}$ with a vanilla loss function $\mathcal{L}_{vmatch}$ that could be applied to $v$.
Recall that $v$ should be similar for inputs from the same TPV and different for inputs from different TPVs.
We define $\mathcal{L}_{vmatch}=\mathcal{L}_{vsim}+\mathcal{L}_{vdissim}$ where $\mathcal{L}_{vsim}=\cos(v_i, v_j)$ and $\mathcal{L}_{vdissim}=\max\{\cos(v_i, v_j), 0\}$ for similarities and dissimilarities between $v$ respectively.
Results show that using $\mathcal{L}_{permute}$ has a lower alignment error (0.1550) than $\mathcal{L}_{vmatch}$ (0.2518), indicating $\mathcal{L}_{permute}$ is reasonable and effective on supervising $v$.
Using $\mathcal{L}_{vmatch}$ leads to a higher alignment error than not using $\mathcal{L}_{permute}$ (0.1922), suggesting that $\mathcal{L}_{vmatch}$ provides a poor supervise signal.
We infer that the $\mathcal{L}_{vdissim}$ part overemphasizes the dissimilarity among $v$ to satisfy $\cos(v_i, v_j) \leq 0$.
We also investigate the performance if we remove certain loss functions.
By removing $\mathcal{L}_{recon}$ and keep $\mathcal{L}_{permute}$ (0.1748), we know that adding a general reconstruction error would help the training of the decoder.
But when we compare using $\mathcal{L}_{tc}$ and $\mathcal{L}_{recon}$ (0.1922) and using $\mathcal{L}_{tc}$ (0.1844), we find that using this $\mathcal{L}_{recon}$ without $\mathcal{L}_{permute}$ is not sufficient to supervise $h$ and $v$.
A well-trained decoder ensures that $\mathcal{L}_{permute}$ is not affected by the under-trained decoder.
As for hyper-parameters $\alpha$ and $\beta$, increasing $\alpha$ does not yield a better result while the result of increasing $\beta$ is comparable.
We always keep $\mathcal{L}_{tc}$, otherwise the model will easily fail to capture temporal information between states.
\begin{figure}[htbp]
    \centering
    \includegraphics[scale=0.34]{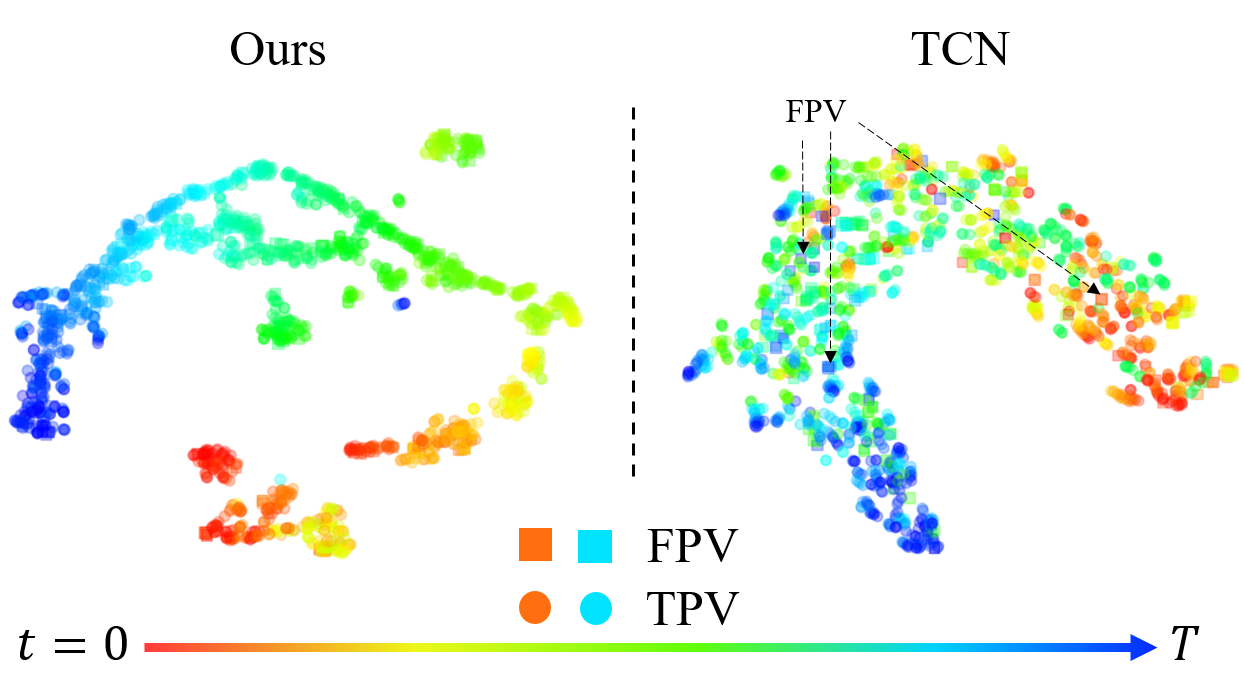}
    \caption{Visualization of state representation spaces $\mathcal{H}$ of Panda environment. Squares are FPV representations and dots are TPV representations. The colors from red to blue indicate the timesteps from 0 to $T$ (end of a trajectory).}
    \label{fig:tsne}
\end{figure}
\subsection{Qualitative Evaluations on Representation Space}
To evaluate the state representation space qualitatively, we first show a t-SNE visualization\cite{van2008visualizing} of the learned representation space on Panda environment (Figure \ref{fig:tsne}).
We observe a generally clear temporal order from both methods.
Moreover, FPV and TPV frames are aligned together by temporal order by our method, which indicates a good joint representation for FPV and TPV. 
However, in the visualization of TCN\cite{sermanet2018time}'s representation space, we observe many FPV representations are scattered in the space and are misaligned.
This will affect the joint representation and generate a misleading reward function in third-person imitation learning.
Therefore our model is more suitable to deal with the visual differences between $\mathcal{O}^F$ and $\mathcal{O}^T$.

\begin{figure}[t]
    \centering
    \includegraphics[scale=0.40]{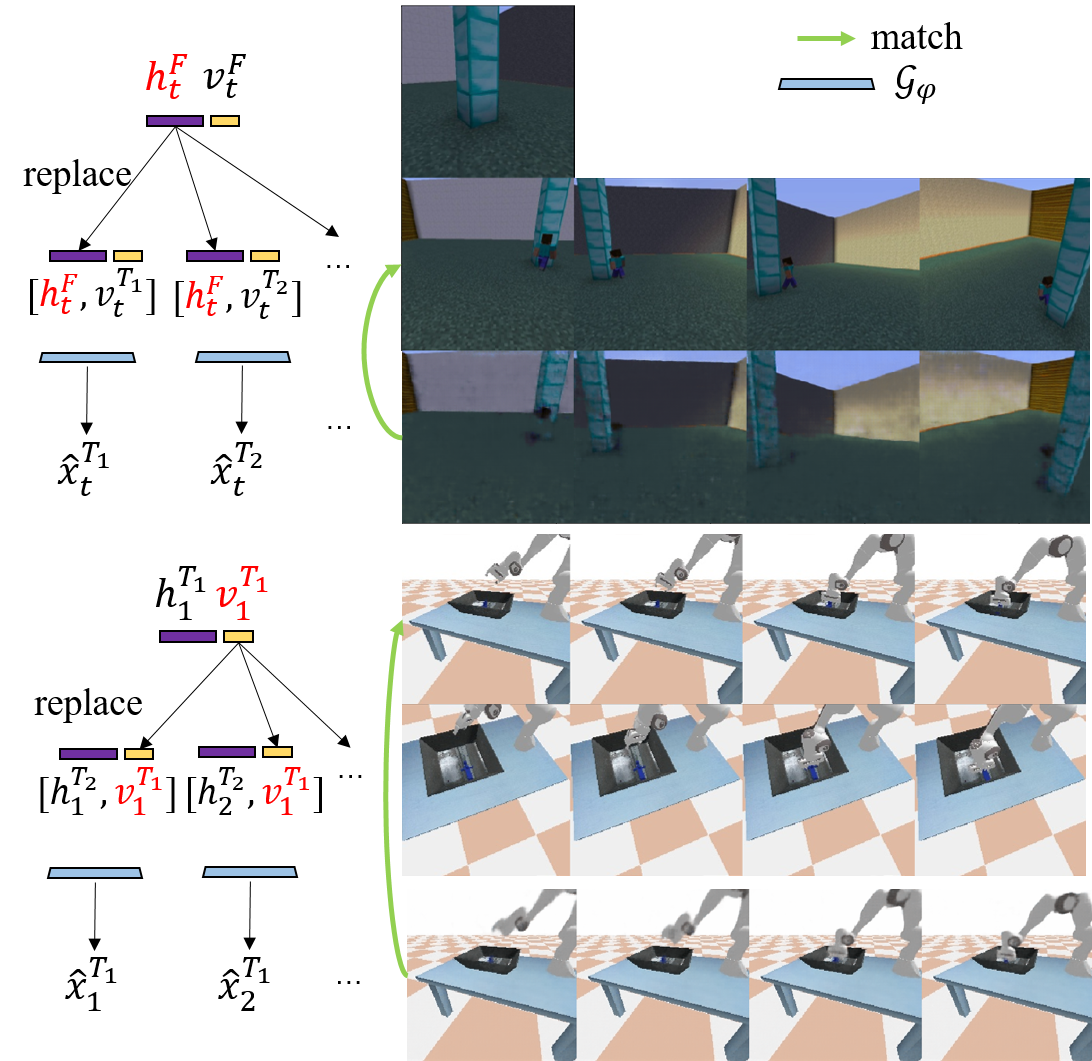}
    \caption{Reconstruction results of representation permutation on the test sets. The green arrow means the reconstructed images (from) should be similar with ground truth (to). Up: replacing $h^{T_j}$ of TPV frames (2nd row) by $h^F$ from an FPV frame (1st row) at the same timestep. Bottom:  replacing $v^{T_j}$ (of 2nd row) by $v^{T_i}$ (of 1st row) from a different viewpoint. }
    \label{fig:vispermute}
    \vspace{-0.1in}
\end{figure}

We then show how our state representation $h$ and viewpoint representation $v$ encode corresponding information i.e. they are well disentangled.
We follow the same permutation and reconstruction procedures as calculating $\mathcal{L}_{permute}$ in Section \ref{sec:permute} on frames in the test set the model has never seen.

Figure \ref{fig:vispermute} shows the result of replacing state representation $h^{T_j}$ with $h^F$ in Minecraft frames and replacing viewpoint representation $v^{T_2}$ with $v^{T_1}$ in Panda frames.
From the result in Panda environment, we could observe the viewpoint in reconstructed images is coherent, which means changing $h$ does not affect the viewpoint.
The states are also well reconstructed if we see the poses of the robot.
In the result of Minecraft environment, we could see the model reconstructs corresponding viewpoints by changing $v$.
Even though the reconstructed agents are blurred, their positions are closed to the ground truth.
We note that reconstruction is not our ultimate goal in this research, but it provides us an informative self-supervised objective to train disentangled latent representations.
The clear disentanglement meets our initial motivation to separate $h$ and $v$ from full representation $z$.

\subsection{Imitation Learning Evaluation}
\begin{table}[t]
    \centering
    \caption{Success rate and cumulative rewards from (single example) imitation learning. *: first-person imitation learning. $^\dagger$: third-person imitation learning.}
    \begin{tabular}{ccc}
    \toprule
    \multirow{3}{*}{\textbf{Model}} & \multicolumn{2}{c}{\textbf{Environment} }  \\
    & Minecraft & Panda    \\
    & \multicolumn{2}{c}{(Succ. Rate / Reward Mean $\pm$ Std)} \\
                              
    \midrule
    Random Policy &  0 / 10.3$\pm$9.47 & 0 / 48.4$\pm$11.0 \\
    \midrule
    Single-view TCN\cite{sermanet2018time} *   & 0.12 / 12.6$\pm$8.33 & 0.16 / 64.0$\pm$8.42\\
    Ours* & 0.46 / 19.8$\pm$9.34 & 0.36 / 69.3$\pm$8.45\\
    
    \midrule
    Multi-view TCN\cite{sermanet2018time}$^\dagger$   & 0.31 / 18.9$\pm$12.4 & 0.32 / 66.4$\pm$9.00\\
    Ours$^\dagger$    & \textbf{0.52} / \textbf{21.9}$\pm$\textbf{10.5} & \textbf{0.44} / \textbf{71.2}$\pm$\textbf{7.01}\\
    \bottomrule
    \end{tabular}
    \label{tab:policy}
    \vspace{-0.1in}
\end{table}
We finally evaluate the quality of our state representation in terms of success rates and rewards by applying it for the actual policy learning.
We try both first-person imitation learning (FPIL) and third-person imitation learning (TPIL) to show that our joint state representation generalizes to both FPV and TPV domains.

\subsubsection{Experiment Settings}
All representation models are trained by expert trajectories in the training set.
We have 6 distinct trajectories for Minecraft environment and 24 for Panda environment.
We implemented two baselines for two different settings: FPIL and TPIL.
For FPIL, we implement a single-view TCN\cite{sermanet2018time} baseline where the representation is trained solely on FPV demonstrations.
For TPIL, we implement a multi-view TCN\cite{sermanet2018time} baseline where the representation is trained by FPV-TPV demonstrations.
Our representation models are trained by FPV-TPV demonstrations as in TCN~\cite{sermanet2018time}, and we apply them to both FPIL and TPIL because our representations are generalizable to both FPV and TPV inputs once learned.
All the policies are trained by providing only one FPV expert demonstration for FPIL, or one expert TPV demonstration for TPIL, from the testing set.

We train each policy on Panda for $10^5$ steps and on Minecraft for $5\times10^4$ steps by PPO\cite{Schulman2017ProximalPO}.
We run 100 evaluation epochs for the best model achieved during training steps.
Successful execution of a task is defined by reaching to the diamond block in Minecraft environment and picking and lifting the object in Panda environment.
A continuous reward is defined by how much distance between the target and the agent has been minimized at any timestep in an epoch.
We regularize the reward according to different randomized target locations to a fair comparison.
The larger reward means a closer approaching to the target for both environments.

\subsubsection{Experiment Results}
Table~\ref{tab:policy} shows that our method outperforms the multi-view TCN\cite{sermanet2018time} baseline in both FPIL and TPIL, suggesting our joint state representation space is better aligned in general and is capable of learning the policy in this egocentric TPIL setting.

In FPIL setting (2nd and 3rd rows in Table~\ref{tab:policy}), our method outperforms the single-view TCN\cite{sermanet2018time} baseline indicating that (1) the representation learned from only few FPV demonstrations is limited and (2) our method takes advantage of the FPV-TPV demonstrations to shape a better state representation space than using pure FPV demonstration.
By introducing third-person demonstrations from multiple viewpoints, we could learn a better representation while keeping the total number of distinct trajectories is small.

\begin{figure}
    \centering
    \includegraphics[width=0.48\textwidth]{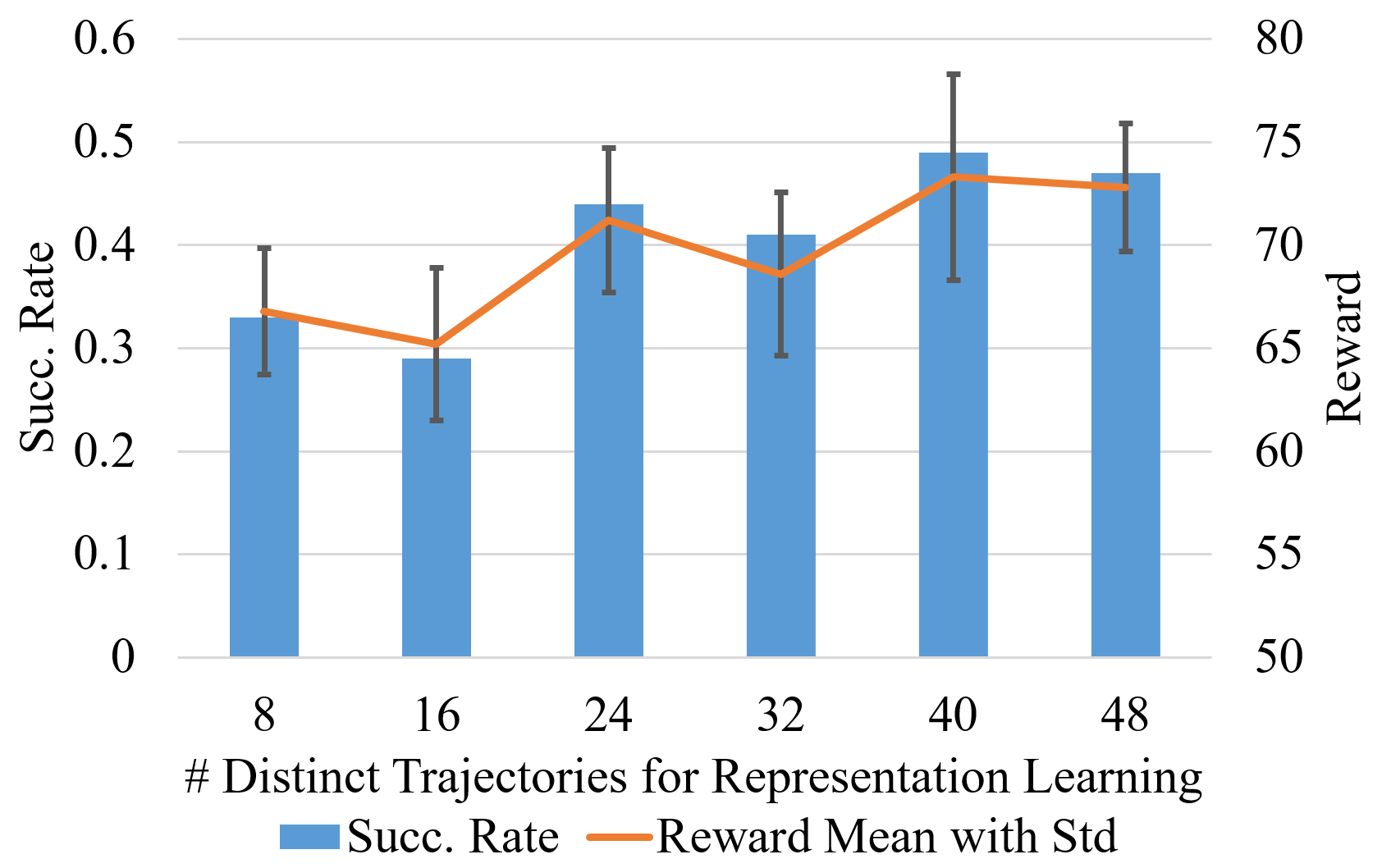}
    \caption{Performance under different number of distinct trajectories used in representation learning. The experiment is taken in Panda environment}
    \label{fig:traj}
    \vspace{-0.1in}
\end{figure}

We also show how the number of trajectories used for representation model training will affect the final policy.
This experiment is done using Panda environment and results are shown in Figure~\ref{fig:traj}. Our method benefits from more trajectories when the trajectory count is small and the performance saturates after providing 40 trajectories. Even with 8 and 16 FPV-TPV trajectories (0.33 and 0.29 Succ. Rate), our method can reach a better performance than Single-view TCN~\cite{sermanet2018time} (0.16), which is trained by 24 FPV trajectories. This highlights the value of our approach, as it uses fewer trajectories to gain better performance.

\section{CONCLUSION}
In this research, we consider a third-person imitation learning setting where the agent performs tasks that actions may cause agents' egomotion.
We solve this egocentric third-person imitation learning by learning a joint state representation space for FPV and TPV inputs.
We introduce a dual AE model to encode FPV and TPV inputs separately considering the visual differences between FPV and TPV videos. 
We explicitly split latent representation into state and viewpoint representations and train them to be disentangled by applying time-contrastive, representation permutation, and reconstruction losses in a self-supervised way.
Results show that our representation space successfully encodes the state information with viewpoint information disentangled.
We apply our joint state representation to both third-person and first-person imitation learning, and results show that our state representation is effective for learning a task out of either TPV or FPV expert demonstrations.
\section*{ACKNOWLEDGMENT}
We thank the reviewers for their comments that greatly improved the manuscript. We would also show our gratitude to Xiang Li and Ryan Burgert for inspiring discussions, figure improvements, and writing.

\bibliographystyle{IEEEtran}
\bibliography{IEEEfull, ref}

\begin{thebibliography}{10}
\providecommand{\url}[1]{#1}
\csname url@rmstyle\endcsname
\providecommand{\newblock}{\relax}
\providecommand{\bibinfo}[2]{#2}
\providecommand\BIBentrySTDinterwordspacing{\spaceskip=0pt\relax}
\providecommand\BIBentryALTinterwordstretchfactor{4}
\providecommand\BIBentryALTinterwordspacing{\spaceskip=\fontdimen2\font plus
\BIBentryALTinterwordstretchfactor\fontdimen3\font minus
  \fontdimen4\font\relax}
\providecommand\BIBforeignlanguage[2]{{%
\expandafter\ifx\csname l@#1\endcsname\relax
\typeout{** WARNING: IEEEtran.bst: No hyphenation pattern has been}%
\typeout{** loaded for the language `#1'. Using the pattern for}%
\typeout{** the default language instead.}%
\else
\language=\csname l@#1\endcsname
\fi
#2}}

\bibitem{hussein2017imitation}
A.~Hussein, M.~M. Gaber, E.~Elyan, and C.~Jayne, ``Imitation learning: A survey
  of learning methods,'' \emph{ACM Computing Surveys (CSUR)}, vol.~50, no.~2,
  pp. 1--35, 2017.

\bibitem{premack1978does}
D.~Premack and G.~Woodruff, ``Does the chimpanzee have a theory of mind?''
  \emph{Behavioral and brain sciences}, vol.~1, no.~4, pp. 515--526, 1978.

\bibitem{meltzoff1988imitation}
A.~N. Meltzoff, ``Imitation of televised models by infants,'' \emph{Child
  development}, vol.~59, no.~5, p. 1221, 1988.

\bibitem{Stadie2017ThirdPersonIL}
B.~C. Stadie, P.~Abbeel, and I.~Sutskever, ``Third-person imitation learning,''
  \emph{ArXiv}, vol. abs/1703.01703, 2017.

\bibitem{sermanet2018time}
P.~Sermanet, C.~Lynch, Y.~Chebotar, J.~Hsu, E.~Jang, S.~Schaal, S.~Levine, and
  G.~Brain, ``Time-contrastive networks: Self-supervised learning from video,''
  in \emph{2018 IEEE International Conference on Robotics and Automation
  (ICRA)}.\hskip 1em plus 0.5em minus 0.4em\relax IEEE, 2018, pp. 1134--1141.

\bibitem{mees2020adversarial}
O.~Mees, M.~Merklinger, G.~Kalweit, and W.~Burgard, ``Adversarial skill
  networks: Unsupervised robot skill learning from video,'' in \emph{2020 IEEE
  International Conference on Robotics and Automation (ICRA)}.\hskip 1em plus
  0.5em minus 0.4em\relax IEEE, 2020, pp. 4188--4194.

\bibitem{edwards2018imitating}
A.~D. Edwards, H.~Sahni, Y.~Schroecker, and C.~L. Isbell, ``Imitating latent
  policies from observation,'' \emph{arXiv preprint arXiv:1805.07914}, 2018.

\bibitem{liu2018imitationfrom}
\BIBentryALTinterwordspacing
Y.~Liu, A.~Gupta, P.~Abbeel, and S.~Levine, ``Imitation from observation:
  Learning to imitate behaviors from raw video via context translation,''
  \emph{2018 IEEE International Conference on Robotics and Automation (ICRA)},
  May 2018. [Online]. Available:
  \url{http://dx.doi.org/10.1109/ICRA.2018.8462901}
\BIBentrySTDinterwordspacing

\bibitem{yang2020cross}
S.~Yang, W.~Zhang, W.~Lu, H.~Wang, and Y.~Li, ``Cross-context visual imitation
  learning from demonstrations,'' in \emph{2020 IEEE International Conference
  on Robotics and Automation (ICRA)}.\hskip 1em plus 0.5em minus 0.4em\relax
  IEEE, 2020, pp. 5467--5473.

\bibitem{ho2016generative}
\BIBentryALTinterwordspacing
J.~Ho and S.~Ermon, ``Generative adversarial imitation learning,'' \emph{CoRR},
  vol. abs/1606.03476, 2016. [Online]. Available:
  \url{http://arxiv.org/abs/1606.03476}
\BIBentrySTDinterwordspacing

\bibitem{aytar2018playing}
Y.~Aytar, T.~Pfaff, D.~Budden, T.~L. Paine, Z.~Wang, and N.~de~Freitas,
  ``Playing hard exploration games by watching youtube,'' 2018.

\bibitem{dwibedi2018learning}
D.~Dwibedi, J.~Tompson, C.~Lynch, and P.~Sermanet, ``Learning actionable
  representations from visual observations,'' in \emph{2018 IEEE/RSJ
  International Conference on Intelligent Robots and Systems (IROS)}.\hskip 1em
  plus 0.5em minus 0.4em\relax IEEE, 2018, pp. 1577--1584.

\bibitem{kingma2013auto}
D.~P. Kingma and M.~Welling, ``Auto-encoding variational bayes,'' \emph{arXiv
  preprint arXiv:1312.6114}, 2013.

\bibitem{goodfellow2014generative}
I.~J. Goodfellow, J.~Pouget-Abadie, M.~Mirza, B.~Xu, D.~Warde-Farley, S.~Ozair,
  A.~C. Courville, and Y.~Bengio, ``Generative adversarial nets,'' in
  \emph{NIPS}, 2014.

\bibitem{chen2018isolating}
R.~T.~Q. Chen, X.~Li, R.~Grosse, and D.~Duvenaud, ``Isolating sources of
  disentanglement in variational autoencoders,'' in \emph{Advances in Neural
  Information Processing Systems}, 2018.

\bibitem{sohn2015learning}
K.~Sohn, X.~Yan, and H.~Lee, ``Learning structured output representation using
  deep conditional generative models,'' in \emph{Proceedings of the 28th
  International Conference on Neural Information Processing Systems-Volume 2},
  2015, pp. 3483--3491.

\bibitem{chen2016infogan}
X.~Chen, Y.~Duan, R.~Houthooft, J.~Schulman, I.~Sutskever, and P.~Abbeel,
  ``Infogan: interpretable representation learning by information maximizing
  generative adversarial nets,'' in \emph{Neural Information Processing Systems
  (NIPS)}, 2016.

\bibitem{karras2019astyle}
\BIBentryALTinterwordspacing
T.~Karras, S.~Laine, and T.~Aila, ``A style-based generator architecture for
  generative adversarial networks,'' \emph{2019 IEEE/CVF Conference on Computer
  Vision and Pattern Recognition (CVPR)}, Jun 2019. [Online]. Available:
  \url{http://dx.doi.org/10.1109/CVPR.2019.00453}
\BIBentrySTDinterwordspacing

\bibitem{nair2020contextual}
A.~Nair, S.~Bahl, A.~Khazatsky, V.~Pong, G.~Berseth, and S.~Levine,
  ``Contextual imagined goals for self-supervised robotic learning,'' in
  \emph{Conference on Robot Learning}.\hskip 1em plus 0.5em minus 0.4em\relax
  PMLR, 2020, pp. 530--539.

\bibitem{zolna2019task}
K.~Zolna, S.~Reed, A.~Novikov, S.~G. Colmenarej, D.~Budden, S.~Cabi, M.~Denil,
  N.~de~Freitas, and Z.~Wang, ``Task-relevant adversarial imitation learning,''
  \emph{arXiv preprint arXiv:1910.01077}, 2019.

\bibitem{peng2019domain}
X.~Peng, Z.~Huang, X.~Sun, and K.~Saenko, ``Domain agnostic learning with
  disentangled representations,'' in \emph{International Conference on Machine
  Learning}.\hskip 1em plus 0.5em minus 0.4em\relax PMLR, 2019, pp. 5102--5112.

\bibitem{jiang2019disentangled}
Z.-H. Jiang, Q.~Wu, K.~Chen, and J.~Zhang, ``Disentangled representation
  learning for 3d face shape,'' in \emph{Proceedings of the IEEE/CVF Conference
  on Computer Vision and Pattern Recognition}, 2019, pp. 11\,957--11\,966.

\bibitem{shu2017NeuralFace}
Z.~Shu, E.~Yumer, S.~Hadap, K.~Sunkavalli, E.~Shechtman, and D.~Samaras,
  ``Neural face editing with intrinsic image disentangling,'' in \emph{Computer
  Vision and Pattern Recognition, 2017. CVPR 2017. IEEE Conference on}.\hskip
  1em plus 0.5em minus 0.4em\relax IEEE, 2017, pp.~--.

\bibitem{zhang2019gait}
Z.~Zhang, L.~Tran, X.~Yin, Y.~Atoum, X.~Liu, J.~Wan, and N.~Wang, ``Gait
  recognition via disentangled representation learning,'' in \emph{Proceedings
  of the IEEE/CVF Conference on Computer Vision and Pattern Recognition}, 2019,
  pp. 4710--4719.

\bibitem{denton2017unsupervised}
E.~Denton and V.~Birodkar, ``Unsupervised learning of disentangled
  representations from video,'' in \emph{Proceedings of the 31st International
  Conference on Neural Information Processing Systems}, 2017, pp. 4417--4426.

\bibitem{aytar2017cross}
Y.~Aytar, L.~Castrejon, C.~Vondrick, H.~Pirsiavash, and A.~Torralba,
  ``Cross-modal scene networks,'' \emph{IEEE transactions on pattern analysis
  and machine intelligence}, vol.~40, no.~10, pp. 2303--2314, 2017.

\bibitem{doersch2015unsupervised}
C.~Doersch, A.~Gupta, and A.~A. Efros, ``Unsupervised visual representation
  learning by context prediction,'' in \emph{Proceedings of the IEEE
  international conference on computer vision}, 2015, pp. 1422--1430.

\bibitem{tian2020cmc}
Y.~Tian, D.~Krishnan, and P.~Isola, ``Contrastive multiview coding,'' in
  \emph{Computer Vision -- ECCV 2020}, A.~Vedaldi, H.~Bischof, T.~Brox, and
  J.-M. Frahm, Eds.\hskip 1em plus 0.5em minus 0.4em\relax Cham: Springer
  International Publishing, 2020, pp. 776--794.

\bibitem{chen2020simple}
T.~Chen, S.~Kornblith, M.~Norouzi, and G.~Hinton, ``A simple framework for
  contrastive learning of visual representations,'' \emph{arXiv preprint
  arXiv:2002.05709}, 2020.

\bibitem{chen2020big}
T.~Chen, S.~Kornblith, K.~Swersky, M.~Norouzi, and G.~Hinton, ``Big
  self-supervised models are strong semi-supervised learners,'' \emph{arXiv
  preprint arXiv:2006.10029}, 2020.

\bibitem{he2020momentum}
\BIBentryALTinterwordspacing
K.~He, H.~Fan, Y.~Wu, S.~Xie, and R.~Girshick, ``Momentum contrast for
  unsupervised visual representation learning,'' \emph{2020 IEEE/CVF Conference
  on Computer Vision and Pattern Recognition (CVPR)}, Jun 2020. [Online].
  Available: \url{http://dx.doi.org/10.1109/cvpr42600.2020.00975}
\BIBentrySTDinterwordspacing

\bibitem{grill2020bootstrap}
J.-B. Grill, F.~Strub, F.~Altché, C.~Tallec, P.~H. Richemond, E.~Buchatskaya,
  C.~Doersch, B.~A. Pires, Z.~D. Guo, M.~G. Azar, B.~Piot, K.~Kavukcuoglu,
  R.~Munos, and M.~Valko, ``Bootstrap your own latent: A new approach to
  self-supervised learning,'' 2020.

\bibitem{chen2020exploring}
X.~Chen and K.~He, ``Exploring simple siamese representation learning,'' 2020.

\bibitem{srinivas2020curl}
A.~Srinivas, M.~Laskin, and P.~Abbeel, ``Curl: Contrastive unsupervised
  representations for reinforcement learning,'' 2020.

\bibitem{hermans2017defense}
A.~Hermans, L.~Beyer, and B.~Leibe, ``In defense of the triplet loss for person
  re-identification,'' \emph{arXiv preprint arXiv:1703.07737}, 2017.

\bibitem{Higgins2017betaVAELB}
I.~Higgins, L.~Matthey, A.~Pal, C.~Burgess, X.~Glorot, M.~Botvinick,
  S.~Mohamed, and A.~Lerchner, ``beta-vae: Learning basic visual concepts with
  a constrained variational framework,'' in \emph{ICLR}, 2017.

\bibitem{andrychowicz2018hindsight}
M.~Andrychowicz, F.~Wolski, A.~Ray, J.~Schneider, R.~Fong, P.~Welinder,
  B.~McGrew, J.~Tobin, P.~Abbeel, and W.~Zaremba, ``Hindsight experience
  replay,'' 2018.

\bibitem{Schulman2017ProximalPO}
J.~Schulman, F.~Wolski, P.~Dhariwal, A.~Radford, and O.~Klimov, ``Proximal
  policy optimization algorithms,'' \emph{ArXiv}, vol. abs/1707.06347, 2017.

\bibitem{mattew2016malmo}
M.~Johnson, K.~Hofmann, T.~Hutton, and D.~Bignell, ``The malmo platform for
  artificial intelligence experimentation,'' in \emph{Proceedings of the
  Twenty-Fifth International Joint Conference on Artificial Intelligence}, ser.
  IJCAI'16.\hskip 1em plus 0.5em minus 0.4em\relax AAAI Press, 2016, p.
  4246–4247.

\bibitem{guss2019minerldata}
\BIBentryALTinterwordspacing
W.~H. Guss, B.~Houghton, N.~Topin, P.~Wang, C.~Codel, M.~Veloso, and
  R.~Salakhutdinov, ``Mine{RL}: A large-scale dataset of {M}inecraft
  demonstrations,'' \emph{Twenty-Eighth International Joint Conference on
  Artificial Intelligence}, 2019. [Online]. Available: \url{http://minerl.io}
\BIBentrySTDinterwordspacing

\bibitem{coumans2019}
E.~Coumans and Y.~Bai, ``Pybullet, a python module for physics simulation for
  games, robotics and machine learning,'' \url{http://pybullet.org},
  2016--2019.

\bibitem{van2008visualizing}
L.~Van~der Maaten and G.~Hinton, ``Visualizing data using t-sne.''
  \emph{Journal of machine learning research}, vol.~9, no.~11, 2008.

\end{thebibliography}

% \begin{thebibliography}{99}
% % \bibitem{c3} H. Poor, An Introduction to Signal Detection and Estimation.   New York: Springer-Verlag, 1985, ch. 4.
% \end{thebibliography}

\end{document}